%% file: main.tex
\PassOptionsToPackage{table,xcdraw}{xcolor}
\documentclass{article}

\usepackage{microtype}
\usepackage{graphicx}
\usepackage{subfigure}
\usepackage{booktabs} 

\usepackage{hyperref}

\usepackage[accepted]{icml2025}

\usepackage{amsmath}
\usepackage{adjustbox}
\usepackage{amssymb}
\usepackage{mathtools}
\usepackage{amsthm}

\newcommand{\model}{\textsc{ViLaMP}}
\usepackage[capitalize,noabbrev]{cleveref}
\usepackage{booktabs}
\usepackage{multirow}

\theoremstyle{plain}

\theoremstyle{definition}

\theoremstyle{remark}

\usepackage{algorithm}
\usepackage[algo2e,ruled,linesnumbered,noline,noend]{algorithm2e}
\newcommand{\tcpo}[1]{\tcp{{\textcolor{blue}{\textsf{#1}}}}}

\usepackage[textsize=tiny]{todonotes}

\begin{document}

\twocolumn[
\icmltitle{Scaling Video-Language Models to 10K Frames via Hierarchical Differential Distillation}


\icmlsetsymbol{equal}{*}
\icmlsetsymbol{cp}{$\dagger$}

\begin{icmlauthorlist}
\icmlauthor{Chuanqi Cheng}{equal,ruc,comp}
\icmlauthor{Jian Guan}{equal,comp}
\icmlauthor{Wei Wu}{cp,comp}
\icmlauthor{Rui Yan}{cp,ruc,wAI}
\end{icmlauthorlist}

\icmlaffiliation{ruc}{Gaoling School of Artificial Intelligence, Renmin University of China}
\icmlaffiliation{comp}{Ant Group}
\icmlaffiliation{wAI}{School of Artificial Intelligence, Wuhan University}

\icmlcorrespondingauthor{Wei Wu, Rui Yan}{ruiyan@ruc.edu.cn}

\icmlkeywords{Machine Learning, ICML}

\vskip 0.3in
]



\printAffiliationsAndNotice{\icmlEqualContribution} 

\begin{abstract}

Long-form video processing fundamentally challenges vision-language models (VLMs) due to the high computational costs of handling extended temporal sequences. Existing token pruning and feature merging methods often sacrifice critical temporal dependencies or dilute semantic information. We introduce differential distillation, a principled approach that systematically preserves task-relevant information while suppressing redundancy. Based on this principle, we develop \model, a hierarchical video-language model that processes hour-long videos at ``mixed precision'' through two key mechanisms: (1) differential keyframe selection that maximizes query relevance while maintaining temporal distinctiveness at the frame level and (2) differential feature merging that preserves query-salient features in non-keyframes at the patch level. Hence, \model~retains full information in keyframes while reducing non-keyframes to their most salient features, resembling mixed-precision training~\cite{micikevicius2018mixed}. Extensive experiments demonstrate \model's superior performance across four video understanding benchmarks, particularly on long-form content. Notably, \model~can process ultra-long videos (up to 10K frames) on a single NVIDIA A100 GPU, achieving substantial computational efficiency while maintaining state-of-the-art performance. 
Code and model are available at \url{https://github.com/steven-ccq/ViLAMP}.

\end{abstract}

\input{introduction}
\input{related_work}
\input{preliminary_study}
\input{method}
\input{experiments}
\input{conclusion}

\section*{Acknowledgments}
This work was supported by the Beijing Outstanding Young Scientist Program NO. BJJWZYJH012019100020098, and Intelligent Social Governance Platform, Major Innovation \& Planning Interdisciplinary Platform for the ``Double-First Class'' Initiative, Renmin University of China, the Fundamental Research Funds for the Central Universities, and Public Computing Cloud, Renmin University of China, the fund for building world-class universities (disciplines) of Renmin University of China.

\section*{Impact Statement}



This paper aims to propose a video-language model that efficiently handles ultra-long video inputs with relatively low computational cost. We would like to highlight a potential impact related to the generated output of our model, which may contain inaccuracies, biases, or offensive content. These issues may stem from various factors, including inherent biases in the training datasets, or the complex nature of video-language interpretation. Such outputs could potentially lead to misinformation propagation or unintended consequences in real-world applications. Therefore, users are expected to implement rigorous validation protocols when deploying this model.

\nocite{langley00}

\bibliography{ref}
\bibliographystyle{icml2025}

\newpage
\appendix
\onecolumn
\input{appendix}



\end{document}

%% file: introduction.tex
\begin{figure}[!t]
    \centering
    \includegraphics[width=0.5\textwidth]{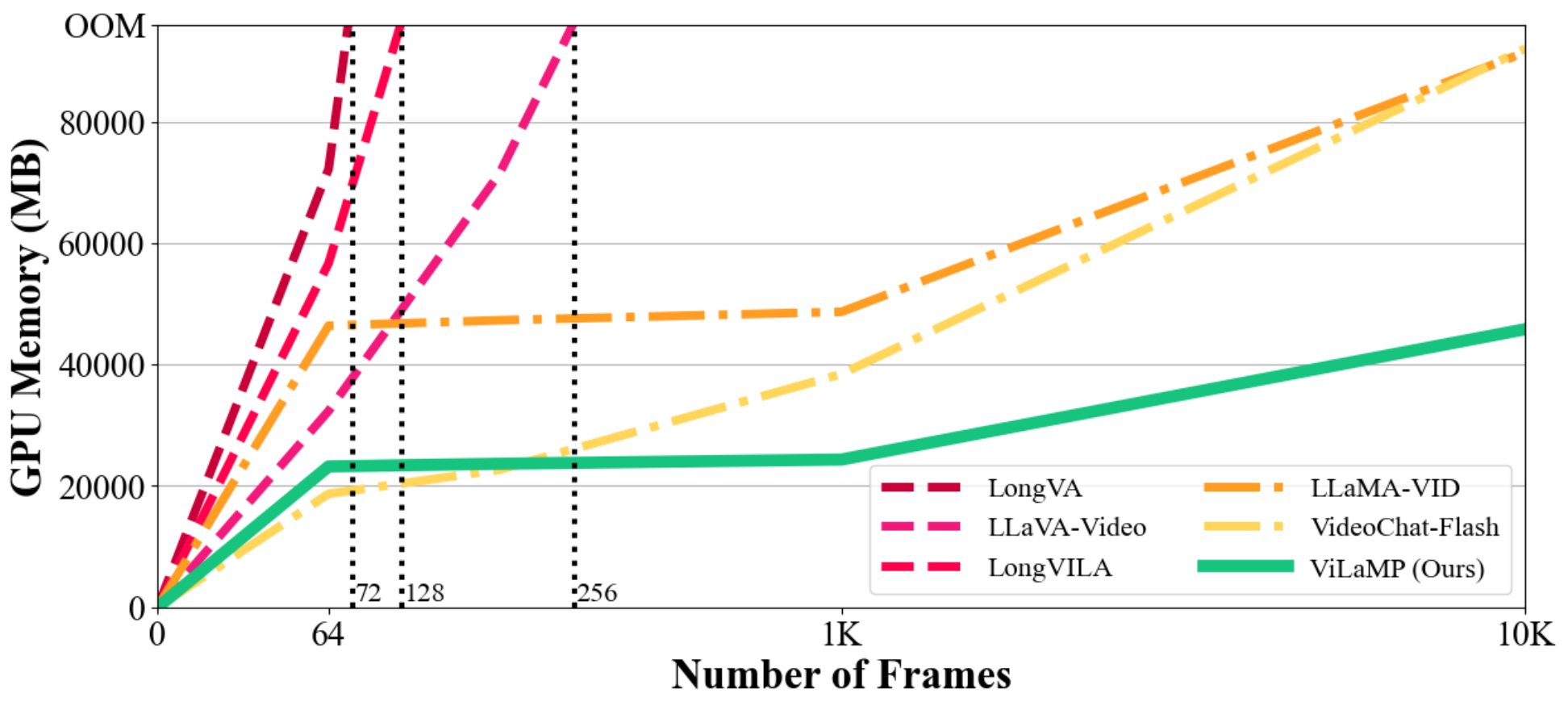}
    \caption{Comparison of the GPU memory consumption with varying input lengths (from 0 to 10K frames) on our VideoNIAH benchmark~(c.f. \S\ref{niah_sec}). OOM indicates out-of-memory errors. All models are evaluated using one NVIDIA A100 GPU.}
    \label{fig:demo}
\end{figure}

\begin{figure*}[!t]
    \centering
    
    \subfigure[LLaMA-VID]{
    \includegraphics[width=.31\textwidth]{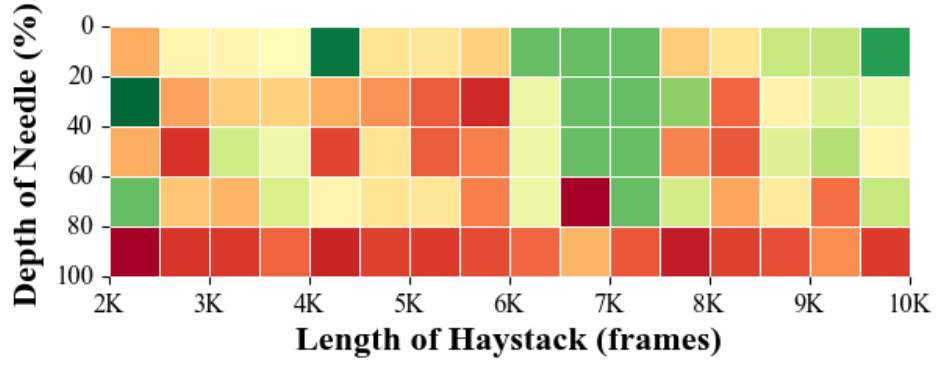}
    }
    \subfigure[VideoChat-Flash]{
    \includegraphics[width=.31\textwidth]{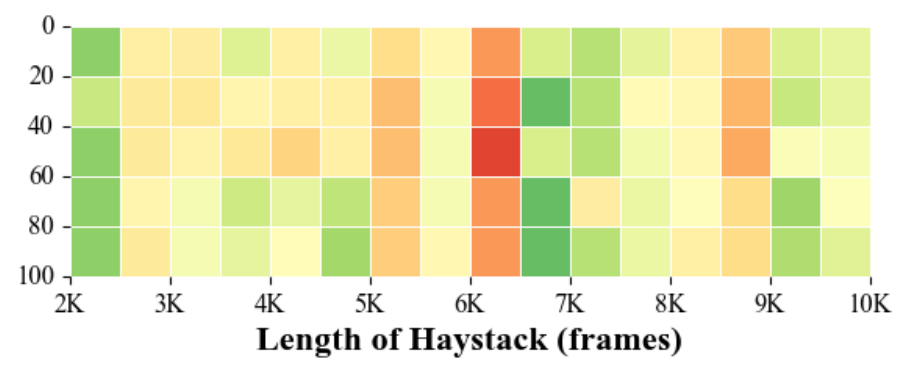}
    }
    \subfigure[\model~(Ours)]{
    \includegraphics[width=.32\textwidth]{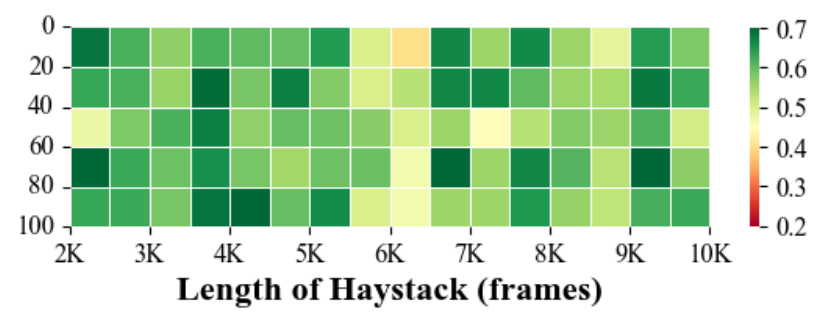}
    }
    \caption{Comparison of ultra-long video understanding capabilities on our VideoNIAH benchmark~(c.f. \S\ref{niah_sec}). Each heatmap shows model performance across different haystack lengths and needle depths. The depth refers to the relative position of the needle video within the haystack with 0\% at the front and 100\% at the end.}
    \label{fig:NIAH}
\end{figure*}

\section{Introduction}
The rapid advancement of vision-language models (VLMs) continues to drive remarkable progress in 
processing both visual and textual content~\cite{openaigpt4o,liu2024visual,chen2024longvilascalinglongcontextvisual,zhang2024longcontexttransferlanguage}. However, as models expand their capabilities from processing single images (typically requiring $729$ visual tokens~\cite{zhai2023sigmoid}) to handling long videos, they inevitably encounter a fundamental challenge: the visual token sequence from a video can easily exceed the context length of LLMs, imposing prohibitive costs in both computational resources and inference latency. For instance, a one-minute video clip at 24 frames per second (FPS) could generate over $1$ million visual tokens ($24 \times 60 \times 729$), far surpassing the context capacity of most mainstream LLMs, which typically ranges from 4K to 128K tokens~\cite{dubey2024llama,yang2024qwen2}. The challenge becomes even more critical as video lengths extend to an hour or beyond, despite the prevalence of such lengthy videos in real-world applications, including long-form video analysis~\cite{lin2023video} and continuous robot learning~\cite{pmlr-v205-wu23c,baker2022video}.




Prior works have explored two main approaches to handle long videos efficiently: 
(1) token pruning through uniform or content-aware sampling strategies~\cite{ye2023mplug,shen2024longvu}, and (2) feature merging via heuristic or learnable mechanisms~\cite{lin2024vila,bai2023qwen}. However, these methods face fundamental limitations: token pruning risks losing critical temporal dependencies, while feature merging often leads to information dilution~\cite{liu2024nvilaefficientfrontiervisual}. The challenge of balancing computational efficiency and semantic preservation thus remains unsolved.


In this work, we explore scaling video-language models to handle significantly longer videos by systematically distilling salient signals into minimal visual tokens without sacrificing information integrity. We begin by addressing a fundamental question: \emph{how to identify truly salient signals from long videos?} While the saliency of visual signals is traditionally measured by query relevance~\cite{li2024llamavid}, videos often contain substantial temporal correlations and repeated content~\cite{shen2024longvu}. This suggests that merely considering query relevance might lead to selecting redundant signals. To validate this hypothesis, we conduct a comprehensive analysis at both frame and patch levels (c.f., \S\ref{preliminary}). Our analysis reveals two key findings: (1) At the frame level, $\sim$90\% of query-induced attention weights concentrate on merely 5\% of frames, with these attention-heavy frames showing significant internal similarities. (2) At the patch level, $\sim$50\% of patches from less-attended frames command 80\% of model attention while exhibiting high visual similarity to those in attention-heavy frames, indicating redundant attention to similar visual patterns. These findings inspire our \textit{differential distillation principle} for efficient video modeling: distilling salient information that exhibits maximal correlation with the query intent while maintaining minimal redundancy with respect to the video's temporal context. 

Building on the principle, we introduce \textbf{\model}, an efficient \textbf{\textsc{Vi}}deo-\textbf{\textsc{La}}nguage model specifically designed to process long videos using ``\textbf{M}ixed \textbf{P}recision'' ---
allocating computational resources according to information saliency. \model~implements this mixed-precision strategy throughout a hierarchical compression framework that operates at two levels: (1) At the frame level, \model{} maintains full visual token representations for salient frames (i.e., keyframes) while applying aggressive compression to intermediate frames (i.e., non-keyframes), reducing each to a single token. Frame saliency is determined through our novel \textit{{\textbf{D}}ifferential {\textbf{K}}eyframe {\textbf{S}}election} (\textsc{\textbf{dks}}) mechanism, which identifies keyframes by maximizing query relevance while ensuring temporal distinctiveness. (2) At the patch level, we introduce the \textit{\textbf{D}ifferential \textbf{F}eature \textbf{M}erging} (\textsc{{\textbf{dfm}}}) strategy for non-keyframe compression, which employs a differentiable learning mechanism to preserve query-salient features while 
suppressing redundant visual information.

We conduct extensive evaluations of \model~across five video understanding benchmarks, 
spanning from minutes to nearly 3 hours. 
\model~outperforms state-of-the-art baselines on five benchmarks. Notably, on the ``Long'' subset of Video-MME, which features an average duration of 2,386 seconds, \model~achieves a substantial 3.5\% and 1.6\% absolute accuracy improvement over the previous top-performing model in non-subtitled and subtitled settings, respectively. 
To rigorously assess VLMs' capability in ultra-long video understanding, we introduce VideoNIAH, a challenging benchmark inspired by the Needle-in-a-Haystack (NIAH) evaluation paradigm from LLM research~\cite{fu2024data,kuratov2024searchneedles11mhaystack}. This benchmark requires models to both locate and comprehend specific needle videos embedded within extensive video sequences (haystacks) while answering targeted queries about these needles. As demonstrated in Figure~\ref{fig:demo} and Figure~\ref{fig:NIAH}, when scaling existing models to process 10K frames (approximately 2.7 hours at 1FPS), they either fail entirely due to memory constraints or suffer severe performance drops compared to the 2K-frame counterpart (e.g., $>$24.50\% accuracy degradation for VideoChat-Flash~\cite{li2024videochat}). In contrast, \model~not only successfully processes videos containing up to 10K frames on a single NVIDIA A100 GPU, but also maintains remarkably stable performance with an 12.82\% absolute improvement over VideoChat-Flash\footnote{Unlike NIAH tasks in LLM research where answers can be directly extracted from the needle text, VideoNIAH requires comprehensive video understanding capabilities to interpret visual-temporal events, making it inherently more challenging to achieve the near-perfect accuracy (e.g., 99\%) observed in LLM evaluations.}. These results convincingly demonstrate that \model~establishes a promising direction for both effective and efficient long-form video understanding.
In summary, our primary contributions are threefold: 

\noindent I. We introduce the differential distillation principle that systematically identifies and preserves task-relevant information while actively suppressing redundancy, establishing a principled approach for efficient long video understanding.

\noindent II. We propose \model, a hierarchical architecture that operationalizes differential distillation through keyframe selection and non-keyframe compression. 
reducing non-keyframes to their most salient features. 

\noindent III. Experiments show that \model~outperforms similar-sized VLMs on four video understanding benchmarks while efficiently scaling to 10K frames on a single GPU, opening new opportunities for processing ultra-long videos.

%% file: related_work.tex
\section{Related Work}
\subsection{Vision-Language Models}

Advances in large language models (LLMs)~\cite{dubey2024llama,yang2024qwen2} has catalyzed the development of VLMs that can process visual and textual inputs, broadly classified into two categories: (1) models jointly trained on visual and textual modalities from scratch, such as Flamingo~\cite{alayrac2022flamingo} and KOSMOS~\cite{peng2023kosmos,peng2023kosmos2groundingmultimodallarge}; and (2) models leveraging pre-trained vision encoders like CLIP~\cite{radford2021learning} or SigLIP~\cite{zhai2023sigmoid}, connected to LLMs through projection modules. This design, pioneered by LLaVA~\cite{liu2024visual} with a simple two-layer MLP connector, has been widely adopted in Qwen-VL~\cite{bai2023qwen}, MiniCPM~\cite{yao2024minicpm}, and InternVL~\cite{chen2024internvl}.

While recent VLMs can handle multiple images~\cite{Qwen2VL,li2024llava} and videos~\cite{zhang2024longva,anonymous2024longvila,zhang2024video}, they face significant limitations: multi-image models ignore temporal dependencies, while video models struggle with computational costs for long videos. This highlights the need for more efficient approaches to video understanding.

\subsection{Long-form Video Understanding}
Current approaches to addressing the computational challenges of long video understanding fall into three categories:

{(1) Context Window Extension.} This line of research focuses on enhancing models' capacity for extended temporal contexts.  \citet{liu2024worldmodelmillionlengthvideo} gradually increases input length during training, while LongVA~\cite{zhang2024long} transfers long-context capabilities from text to visual understanding. However, these approaches still face efficiency limitations with extensive videos.

{(2) Token Pruning.} Various pruning strategies have emerged to reduce computational burden. Early models~\cite{lin2023video,ye2024mplug,cheng2024videollama} used uniform frame sampling, i.e., selecting fixed-interval frames. Recent works propose more sophisticated content-aware pruning approaches: Goldfish~\cite{ataallah2024goldfish} employs database-inspired retrieval mechanisms, MovieChat~\cite{song2024moviechat} implements dynamic memory management, and LongVU~\cite{shen2024longvu} introduces content-aware frame grouping and similarity-based patch selection. However, such pruning risks significant information loss.


{(3) Feature Merging.} Recent research has explored merging video features at various representational levels for video compression, through either heuristic methods (e.g., downsampling~\cite{lin2024vila}, pooling~\cite{xu2024pllava,zhang2024video,li2025llama}, similarity-guided merging~\cite{bolya2023token,jin2024chat,li2024videochat}) or learnable approaches (e.g., Q-Former~\cite{li2023blip, bai2023qwen, zhang2025llavaminiefficientimagevideo}, Perceiver Resampler~\cite{yao2024minicpm}). However, these strategies often struggle to maintain semantic fidelity~\cite{liu2024nvilaefficientfrontiervisual}.



In contrast, \model{} preserves salient information through differential distillation, achieving both computational efficiency and information preservation.

%% file: preliminary_study.tex
\section{Preliminary Studies}\label{preliminary}

Video understanding tasks process a sequence of frames $ V = \langle f_1, f_2, \cdots, f_N \rangle $\footnote{Our study focuses on visual information, excluding subtitles or audio tracks.} to answer a query $Q$. Each frame $ f_n $ is divided into $M$ patches: $ f_n = \langle p_n^1, p_n^2, \cdots, p_n^M \rangle $, where each patch $p_n^m$ is encoded into a unique visual token $t_n^m$ through a vision encoder. Consequently, a video sequence generates $MN$ visual tokens in total. This section analyzes information distribution patterns at both frame level (\S\ref{preliminary_fram}) and patch level (\S\ref{preliminary_patch}), leading to our differential distillation principle~(\S\ref{principle}).

\subsection{Frame-Level Redundancy Analysis}\label{preliminary_fram}

We analyze attention patterns of several representative VLMs, including LLaVA-OneVision~\cite{li2024llava}, LLaVA-Video~\cite{zhang2024video}, Qwen2-VL~\cite{Qwen2VL} and LongVA~\cite{zhang2024longva}, on Video-MME~\cite{fu2024video} by sampling $N$=128 frames per video. For each frame $f_n$, we compute its receiving attention weight $a_n$ by aggregating cross-attention weights between query $Q$ and visual tokens $\{t_n^m\}_{m=1}^M$: $a_n=\sum_{m=1}^Ma_n^m\in[0,1]$, The token-level attention weight $a_n^m\in[0,1]$ is obtained by averaging across all query tokens, all layers, and all attention heads.


\begin{figure}[!t]
    \centering
    \small
    
    \subfigure[LLaVA-OneVision]{
    \includegraphics[width=.22\textwidth]{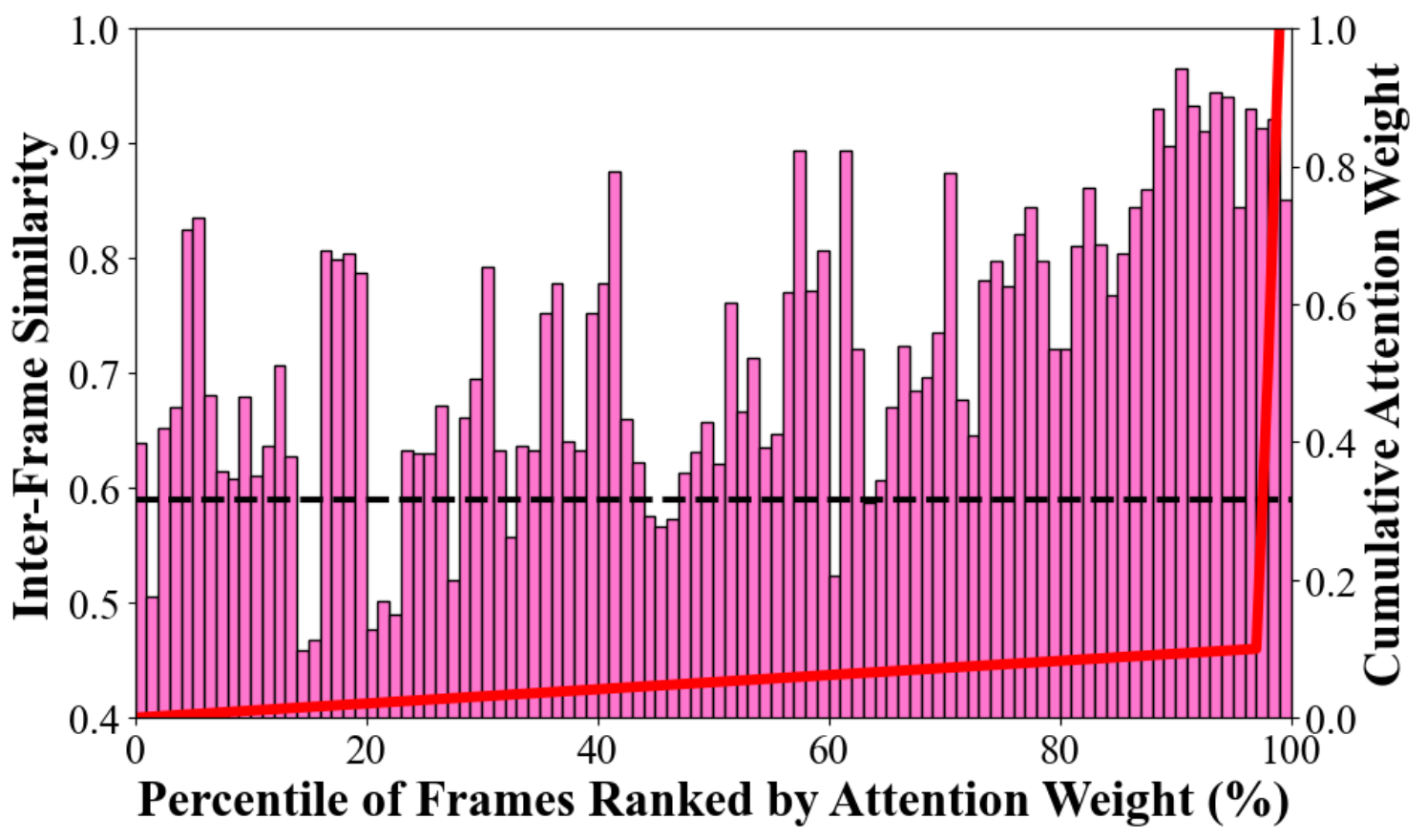}
    }
    \subfigure[LLaVA-Video]{
    \includegraphics[width=.22\textwidth]{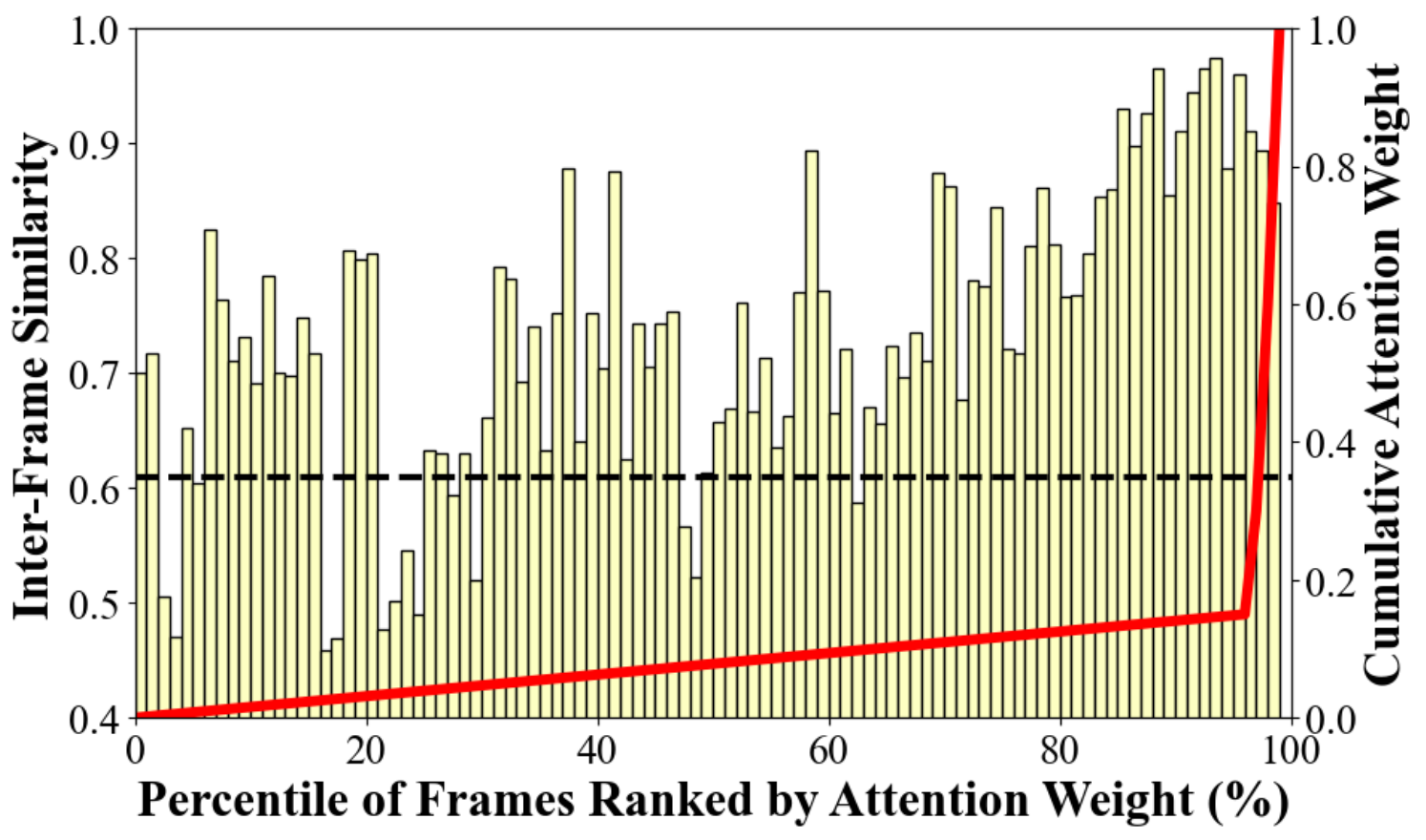}
    }
    \subfigure[Qwen2-VL]{
    \includegraphics[width=.22\textwidth]{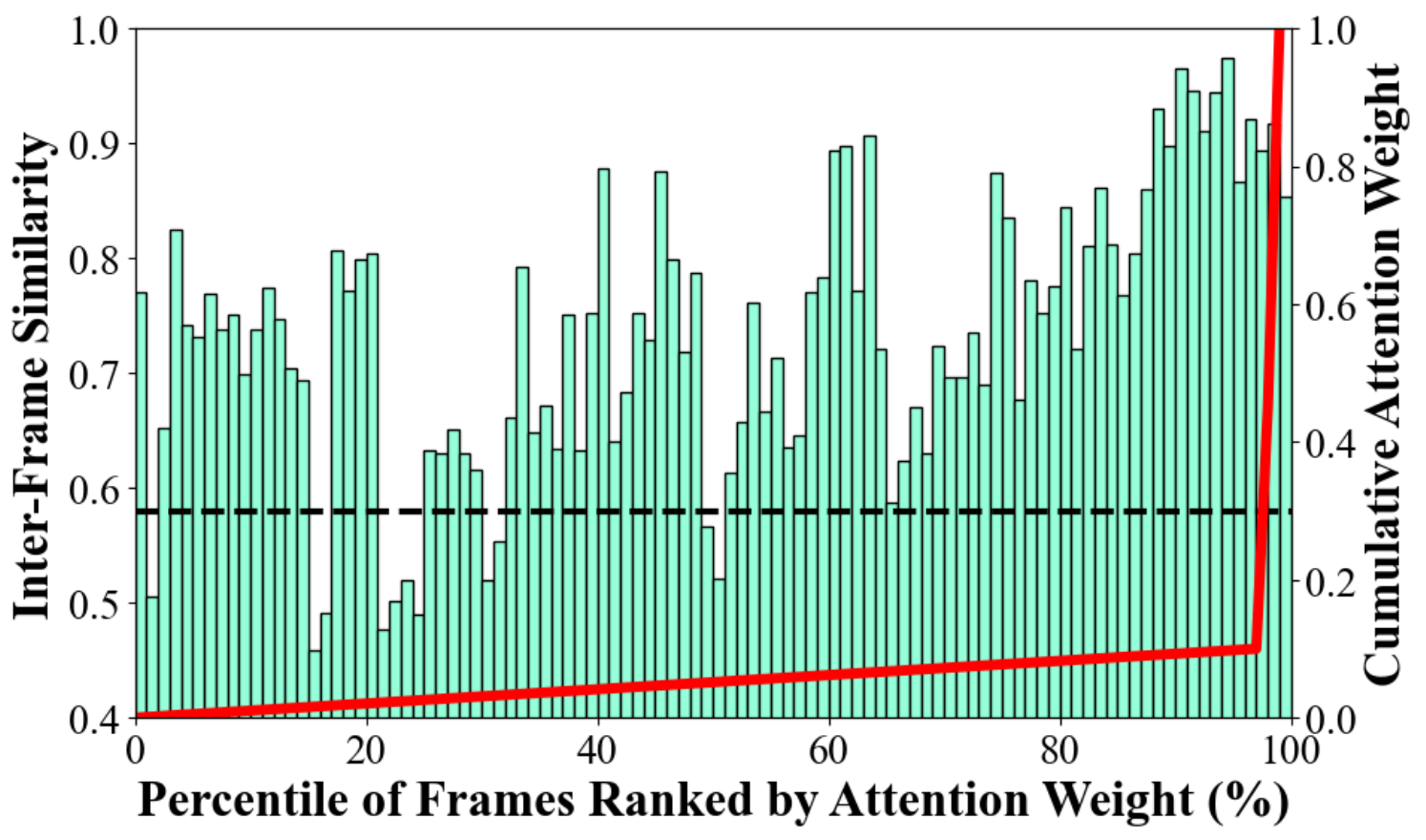}
    }
    \subfigure[LongVA]{
    \includegraphics[width=.22\textwidth]{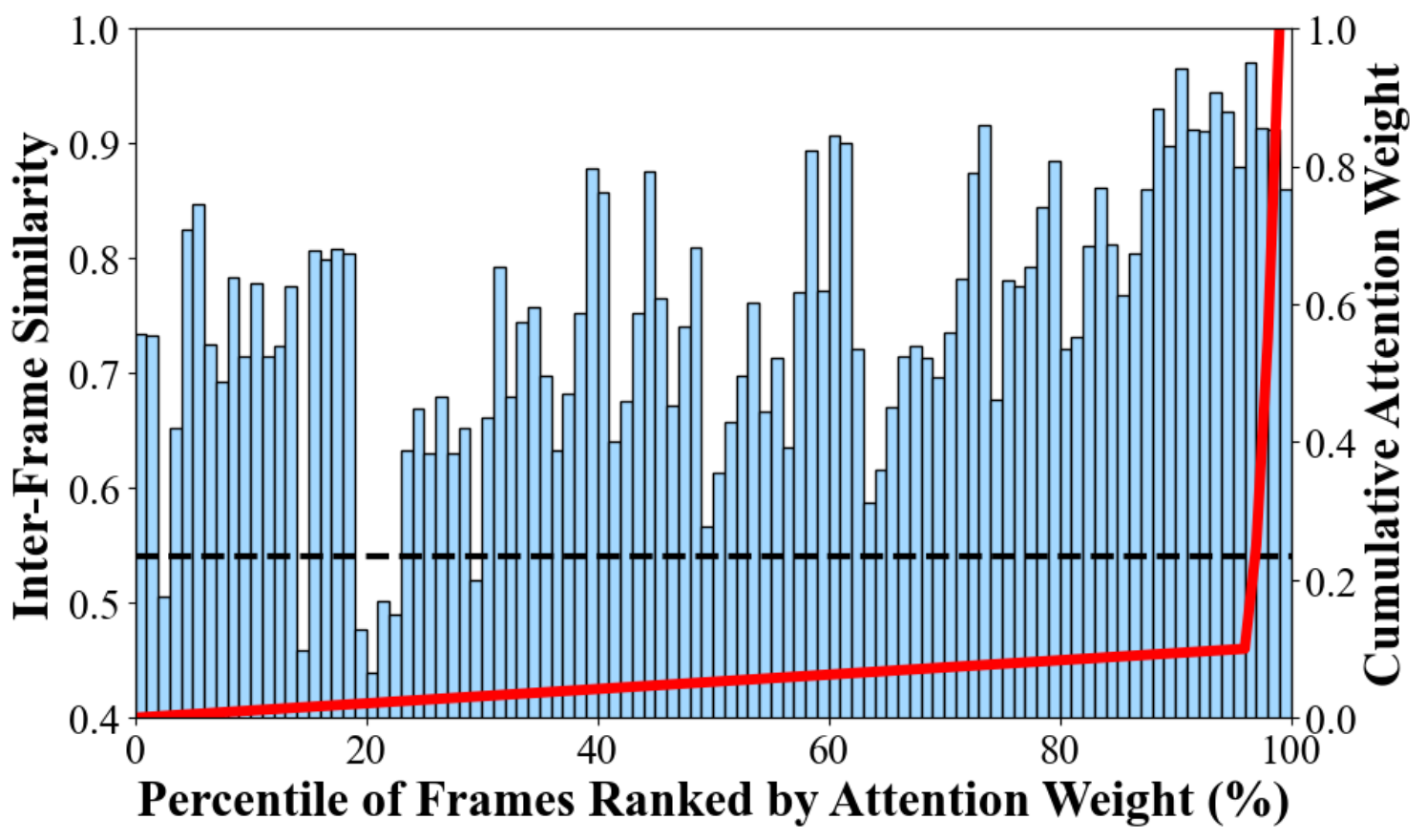}
    }
    \caption{Frame-level attention distribution and frame similarities. $x$-axis: percentile of frames ranked by attention weights from the query (\%). Red curves: cumulative attention. Bars: average similarity with neighboring three frames of similar attention. Dashed lines: random frame pair similarity baseline (0.54$\sim$0.61).}
    \label{fig:visual_token_attention}
\end{figure}


Figure~\ref{fig:visual_token_attention} reveals two key patterns. First, about 90\% of attention mass concentrates on less than 5\% of frames, as evidenced by the steep rise in the cumulative attention curves (red lines). This pronounced attention skew suggests \textbf{query-relevant information clusters around specific keyframes}. Second, and more intriguingly, these attention-heavy frames show high visual similarity (cosine similarity$>$0.8) within themselves, substantially exceeding the random baseline similarity (dashed lines). This phenomenon indicates that \textbf{frames receiving high attention weights contain significant visual redundancy}. Together, these findings provide compelling empirical evidence for developing more efficient video processing strategies by eliminating redundant computation on visually similar, high-attention frames.

\subsection{Patch-Level Redundancy Analysis}
\label{preliminary_patch}
While frame-level redundancy analysis suggests processing only keyframes~\cite{ye2023mplug,shen2024longvu} for efficiency, this approach faces two critical limitations: (1) potential loss of pivotal moments due to selection uncertainty, and (2) missing temporal context essential for tasks like action recognition and causal reasoning. Although this necessitates effective compression strategies for non-keyframes, naive feature merging approaches (e.g., pooling~\cite{xu2024pllava}) can be suboptimal due to substantial information overlap between non-keyframes and keyframes, risking attention misattribution to redundant information.

\begin{figure}[!t]
    \centering
    \includegraphics[width=0.35\textwidth]{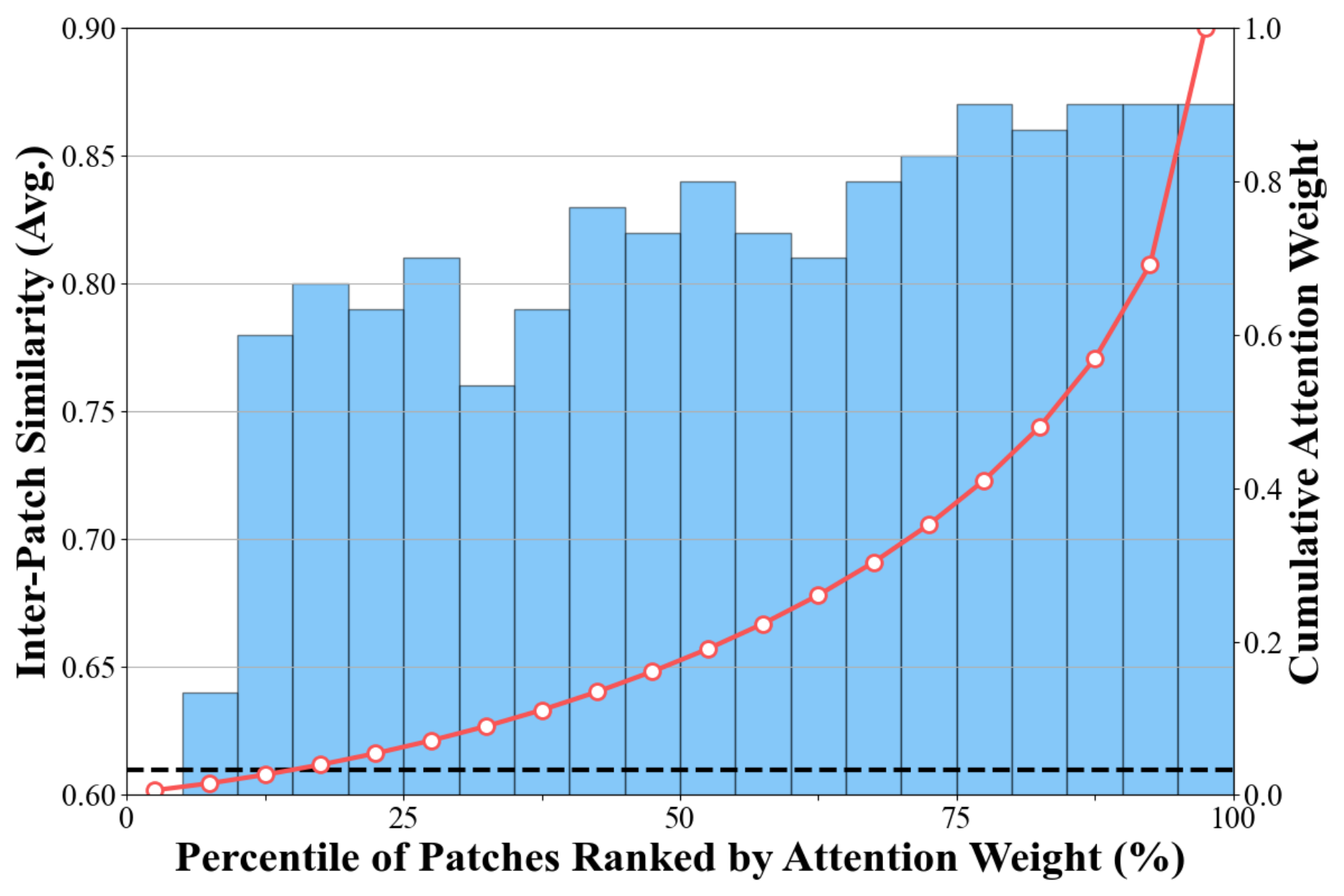}
    \caption{Patch-level attention distribution of LLaVA-OneVision and cross-frame patch similarity. $x$-axis: percentile of non-keyframe patches ranked by attention weights from the query (\%). Red curve: cumulative attention (normalized across all non-keyframe patches). Bars: similarity to nearest keyframe patches. Dashed line: random inter-patch similarity baseline (0.61).}
    
    \label{fig:patch_reduction}
\end{figure}

To verify the hypothesis, we analyze video frames by selecting the top 32 attention-weighted frames as keyframes. For each patch in the remaining 96 non-keyframes, we compute its query attention weight and visual similarity to its corresponding patch in the preceding keyframe (or the first keyframe for earlier non-keyframes). The visual similarity is measured using cosine similarity between the model's visual embeddings. Figure~\ref{fig:patch_reduction} reveals that high-attention patches show strong visual similarity to their keyframe counterparts, with just 50\% of patches contributing 80\% of total attention. This correlation between attention weights and visual similarity indicates that \textbf{substantial computation is spent processing redundant visual patterns across frames.}


\subsection{Differential Distillation Principle}\label{principle}

Our frame- and patch-level analyses motivate our differential distillation principle, fundamentally rethinking how to identify and preserve salient information while reducing redundant computation. The key insight is that truly important information should simultaneously satisfy two criteria: (1) high relevance to the query intent, and (2) low redundancy with respect to its temporal context.  Formally, for any video component $v$ (frame, patch, or feature) and query $Q$, we define its differential information saliency score: \begin{align}
    D(v) = R(v, Q) - T(v, \mathcal{C}(v)),\label{dd_eq}
\end{align}
where $R(v, Q)$ measures query relevance and $T(v, \mathcal{C}(v))$ captures temporal redundancy with context features $\mathcal{C}(v)$. A higher $D(v)$ indicates more unique, task-relevant information deserving more computational resources.

This formulation provides two key advantages: (1) natural adaptation to varying temporal granularities, enabling unified frame- and patch-level operations, and (2) query-aware processing that dynamically allocates computation based on task requirements.

%% file: method.tex
\section{Methodology}

\begin{figure*}[!th]
    \centering
    \includegraphics[width=1.0\textwidth]{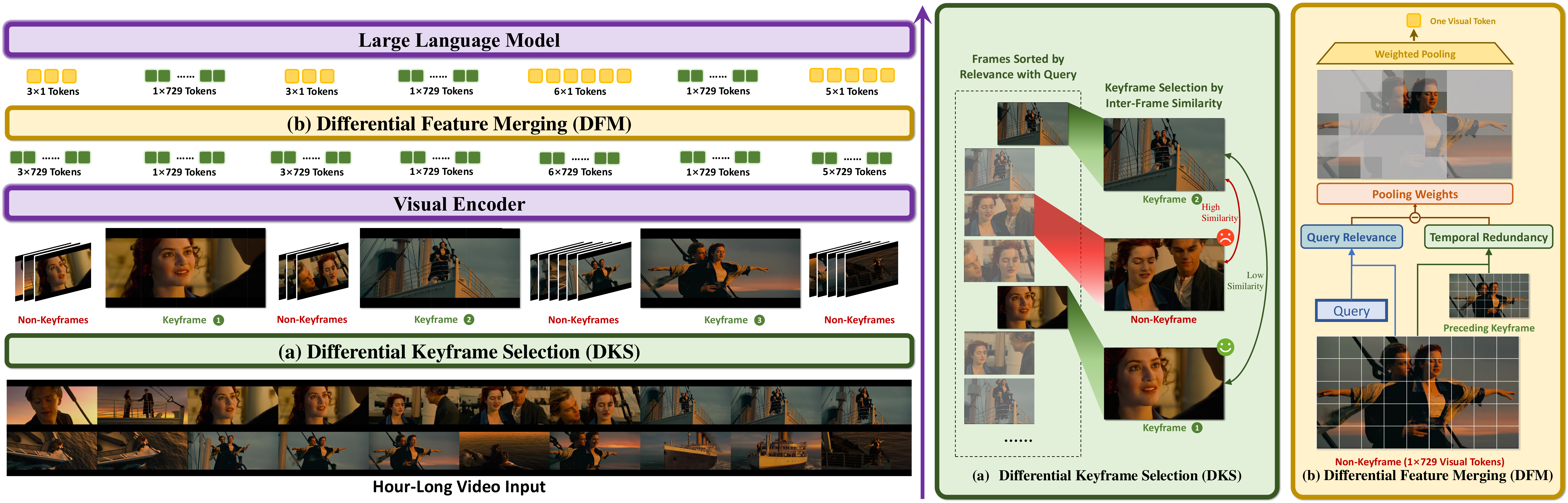}
    \caption{Overview of \model, which first identifies representative keyframes from videos using the differential keyframe selection algorithm (a) and then adaptively merges visual features from non-keyframes through the differential feature merging mechanism (b).}
    \label{fig:overview}
\end{figure*}

Following the differential distillation principle, we propose \model, a hierarchical architecture for mixed-precision video processing, as shown in Figure~\ref{fig:overview}. \model~combines frame-level selection via \textit{Differential Keyframe Selector} (\textsc{dks}) (\S\ref{dks}) and patch-level compression via \textit{Differential Feature Merger} (\textsc{dfm}) (\S\ref{dfm}), enabling efficient resource allocation based on information saliency.


\subsection{Differential Keyframe Selection}\label{dks}


The differential keyframe selection mechanism identifies frames with high query relevance while maintaining temporal distinctiveness. Given frames $V = \langle f_1, f_2, \cdots, f_N \rangle$ and query $Q$, we first encode both the query and each frame into normalized $d$-dimensional embeddings using a CLIP encoder $E_f(\cdot)$~\cite{radford2021clip}, and then compute the frame-level query relevance score $R_f(f_n, Q)$ as the cosine similarity between these embeddings:
\begin{align}
    R_f(f_n, Q) &= \cos(\boldsymbol{f}_n, \boldsymbol{q}),\label{dks_relevance}\\
    \boldsymbol{f}_n=E_f(f_n)\in\mathbb{R}^d,&~~~~\boldsymbol{q}=E_f(Q)\in\mathbb{R}^d,\label{clip}
\end{align}
Then we instantiate the temporal redundancy function in Eq.~\ref{dd_eq} at the frame level as the maximum cosine similarity between frame $f_n$ and its context frames $\mathcal{C}(f_n)$:
\begin{equation}
    T_f(f_n, \mathcal{C}(f_n)) = \max_{f\in\mathcal{C}(f_n)} \cos(\boldsymbol{f}_n, E_f{(f)})
\end{equation}
While one could naively set $\mathcal{C}(f_n)=V\setminus\{f_n\}$, i.e., measuring frame-level redundancy by comparing each frame against all others, this would incur prohibitive quadratic complexity $O(N^2)$. Moreover, selecting frames that directly maximize $R_f-T_f$ without proper prioritization of two terms risks selecting frames that are temporally distinct but irrelevant to the query intent. Therefore, we propose an efficient greedy algorithm (Algorithm~\ref{algorithm1}). It ranks frames by query relevance and selects those maintaining sufficient temporal distance (threshold $\tau$), achieving $O(\text{max}(NK, N\log N))$ complexity for $K$ keyframes while ensuring both semantic relevance and temporal diversity.


\setcounter{AlgoLine}{0}
\begin{algorithm}
\footnotesize
\caption{Differential Keyframe Selection}\label{algorithm1}
\KwData{$V=\langle f_1, f_2, \cdots, f_N \rangle$: Video frames; $Q$: Query; $K$: Maximum keyframes; $\tau$: Similarity threshold.}
\KwResult{$\mathcal{K}$: Keyframe Set.}
\tcpo{Sort all frames by query-relevance in descending order}
$\langle \hat{f}_1, \hat{f}_2, \cdots, \hat{f}_N \rangle=\text{sorted}\big(V, \text{key}=\text{lambda}~f_n: R_f(f_n,Q)\big)$\par
\tcpo{Initialize $\mathcal{K}$ with the most relevant frame}
$\mathcal{K}=\{\hat{f}_1\}$\par
\tcpo{Iterate through the sorted frames}
\While{$n \leftarrow 1$ \KwTo $N$}{
    \If{$|\mathcal{K}| < K$ \textbf{and} $T_f(\hat{f}_n, \mathcal{K})<\tau$}{
        $\mathcal{K}.$add$(\hat{f}_n)$
    }
}
{\Return $\mathcal{K}$}
\end{algorithm}



\subsection{Differential Feature Merging}\label{dfm}

While \textsc{dks} identifies salient keyframes, non-keyframes provide crucial temporal context. Instead of discarding these frames as in mainstream approaches~\cite{yang2024qwen2,chen2024internvl}, we propose a differential feature merging (\textsc{dfm}) approach that compresses non-keyframes to single tokens while preserving task-relevant information.




Specifally, for a non-keyframe $f_n = \langle p_n^1, p_n^2, \cdots, p_n^M \rangle$ and its nearest preceding keyframe $f_k = \langle p_k^1, p_k^2, \cdots, p_k^M \rangle$, we compute patch-level embeddings using the VLM's visual encoder $E_p(\cdot)$, that generates a $d$-dimensional embedding for each patch. The differential information saliency $D_p$ of each patch $p_n^m$ is determined by two factors, i.e., its query relevance $R_p$ and temporal redundancy $T_p$:
\begin{align}
    D_p(p_n^m) &= R_p(p_n^m, Q) - \lambda T_p(p_n^m, p_k^m),\label{lambda_eq}\\
    R_p(p_n^m, Q) &= \cos(\boldsymbol{p}_n^m, \boldsymbol{q}),\\
    T_p(p_n^m, \mathcal{C}(p_n^m)) &= 
    \cos(\boldsymbol{p}_n^m, \boldsymbol{p}_k^m),\\
    \{\boldsymbol{p}_n^m\}_{m=1}^M &= E_p(f_n),~~~ \{\boldsymbol{p}_k^m\}_{m=1}^M = E_p(f_k)
\end{align}
where $m$ is the patch index, $k$/$n$ refers to the keyframe/non-keyframe index,  $\lambda$ balances the trade-off between query relevance and temporal uniqueness, and $\boldsymbol{p}_n^m/\boldsymbol{p}_k^m$ and $\boldsymbol{q}$ represent the encoded patch and query embeddings, respectively. $\mathcal{C}(p_n^m)$ is set to $\{p_k^m\}$ to compute temporal redundancy only with respect to the corresponding patch position in the keyframe, leveraging the strong spatial correspondence prior in video content while maintaining computational efficiency. We set $T_p=0$ for non-keyframes without preceding keyframes.

The saliency scores guide our adaptive token merging strategy through a differentially weighted pooling operation:
\begin{align}
    \boldsymbol{t}_n &= \frac{\sum_{m=1}^M w_n^m \boldsymbol{p}_n^m}{\sum_{m=1}^M w_n^m},\\
    w_n^m &= \text{softmax}(\frac1\alpha [D_p(p_n^1),\cdots, D_p(p_n^M)])|_m,\label{alpha_eq}
\end{align}
where $\boldsymbol{t}_n$ is the compressed representation for the non-keyframe $f_n$, $w_n^m$ is the pooling weights, and $\alpha$ controls the sharpness of the weight distribution. This formulation ensures that patches receive high-importance weights when they contain query-relevant information that is not already captured by the corresponding keyframe patches. 




\subsection{Multimodal Learning}
We utilize a two-stream visual connector architecture to integrate heterogeneous representations from keyframes and compressed non-keyframes into the language model:
\begin{align}
    \boldsymbol{h}_k^m = \text{MLP}_k(\boldsymbol{p}_k^m), ~~~~\boldsymbol{h}_n = \text{MLP}_n(\boldsymbol{t}_n),
\end{align}
where $\boldsymbol{h}_k^m$ is the projected embedding for the $m$-th patch of keyframe $f_k$ and $\boldsymbol{h}_n$ is the embedding for the compressed representation of non-keyframe $f_n$, processed through separate learnable two-layer MLPs. The model is then trained using a language modeling objective over the concatenated sequence of visual embeddings and the prompt:
\begin{align}
    \mathcal{L}=-\log P(A|\{\boldsymbol{h}_k^m|f_k\in\mathcal{K}\}\cup\{\boldsymbol{h}_n|f_n\not\in\mathcal{K}\}, Q),
\end{align}
where the keyframe and non-keyframe embeddings are arranged in their temporal order, $Q$ and $A$ represent the query and target answer, respectively.


This design of \model~offers three key advantages. First, it dramatically reduces the total number of visual tokens from $MN$ to $MK + (N-K)$ where $K\ll N$, enabling efficient processing of extended video sequences. Second, the full-token keyframe representations serve as natural anchors for learning temporal relationships, facilitating the model's interpretation of compressed intermediate frames. Third, the end-to-end optimization of \textsc{dfm} parameters through direct supervision ensures that our compression mechanism adapts to preserve task-relevant information effectively.





%% file: experiments.tex
\section{Experiments}
\input{tables/main_results}
\subsection{Experimental Setup}
\textbf{Benchmarks.} We thoroughly evaluate \model~on four video understanding benchmarks spanning diverse temporal scales and tasks (See Appendix~\ref{appendix:benchmarks} for details): LVBench~\cite{wang2024lvbench} for long-term decision-making, EgoSchema~\cite{mangalam2023egoschema} for natural scenario understanding, LongVideoBench~\cite{wu2024longvideobench} for referred reasoning, and Video-MME~\cite{fu2024video} for comprehensive video understanding.

\textbf{Baselines.} We conduct comprehensive comparisons with state-of-the-art baselines spanning proprietary VLMs, open-source multi-image VLMs, and open-source video-language models. Please refer to Appendix~\ref{appendix:baselines} for more details.


\textbf{Implementation Details.} \model~employs the same architecture as LLaVA-OneVision~\cite{li2024llavaonevisioneasyvisualtask}, utilizing SigLIP-so400m\footnote{\url{https://huggingface.co/google/siglip-so400m-patch14-384}} as the visual encoder, a two-layer MLP as the vision-language connector and Qwen2-7B\footnote{\url{https://huggingface.co/Qwen/Qwen2-7B}} as the language model. We process frames at 384$\times$384 resolution and empirically set $\tau$ in Alg.~\ref{algorithm1} to 0.85, the keyframe count $K$ to 32, $\lambda$ in Eq.~\ref{lambda_eq} to $1$, and $\alpha$ in Eq.~\ref{alpha_eq} to $10^{-2}$ unless otherwise specified. We employ CLIP-ViT-B-32\footnote{\url{https://huggingface.co/sentence-transformers/clip-ViT-B-32}} as $E_f(\cdot)$ in Eq.~\ref{clip}. 
More training details are provided in Appendix~\ref{appendix:training}.


\subsection{Main Results}



Table~\ref{table:main_result} presents comprehensive evaluation results across five benchmarks. Among open-source models of similar size, \model~establishes new state-of-the-art performance across diverse temporal scales, from short-form (EgoSchema) to medium-form (LongVideoBench) and long-form (LVBench and Video-MME) video understanding tasks. Remarkably, \model~performs competitively with or surpasses substantially larger open-source models ($\sim$70B parameters) on these benchmarks. 
On the long-video subset of Video-MME, \model~achieves 67.7\% in non-subtitled setting and 71.6\% in subtitled setting, outperforming existing similar-sized models by a large margin. These consistent strong results across varying temporal scales, particularly in challenging long-form scenarios, validate \model's robust video modeling capabilities.

\subsection{Scaling Up to 10K Frames}\label{niah_sec}

The ability to process hour-long videos is crucial for real-world applications. However, evaluating models' long video understanding capabilities remains challenging. While existing works often adapt the Needle-in-a-Haystack (NIAH) paradigm from LLM research to assess VLMs~\cite{zhang2024long,shen2024longvu}, these adaptations typically exhibit two key limitations: (1) using simplified single-frame ``needles'' that fail to capture temporal dynamics, and (2) limiting haystack sizes to at most 3K frames, inadequately testing true long-form processing capabilities.


To rigorously address above limitations, we develop VideoNIAH, a more challenging variant of the NIAH task specifically designed for video content. We construct VideoNIAH by sampling long-form videos from Video-MME to create ``haystacks'' ranging from 2K to 10K frames (at 1FPS). We then insert ``needle'' video clips (30$\sim$120 seconds) within these haystacks at random positions. Both the needle videos and their corresponding query-answer pairs are sampled from the original Video-MME dataset. Models must not only locate these needles but also comprehend their temporal content to answer targeted queries. We create 3K test cases that  distributed across five haystack lengths (2K, 4K, 6K, 8K, and 10K frames), with carefully balanced question types to ensure comprehensive evaluation. The testing was conducted in a non-subtitled setting.

\textbf{Efficacy.} Figure~\ref{fig:NIAH} compares \model~against LLaMA-VID~\cite{li2024llamavid} and VideoChat-Flash~\cite{li2024videochat} across varying haystack lengths, two state-of-the-art models specifically designed for ultra-long video understanding. The results reveal several key insights: (1) While all models show some performance degradation as context length increases, \model~maintains significantly higher average accuracy at 10K frames (58.15\% vs. 47.25\% of VideoChat-Flash). (2) \model~exhibits remarkably stable performance scaling, with 12.82\% less drop of average accuracy from 2K to 10K frames compared to VideoChat-Flash, 
suggesting superior scalability to ultra-long videos. 

\textbf{Efficiency.} Figure~\ref{fig:demo} shows that \model~exhibits minimal memory growth with increased input length, achieving $\sim$50\% lower memory consumption than baselines when processing 10K frames.
To comprehensively evaluate computational efficiency, we additionally measure the required computational load (FLOPs) and inference latency (Time-to-First-Token, TTFT) under varying input lengths.
As reported in Appendix~\ref{appendix:efficiency}, \model~shows substantial computational efficiency, particularly for longer inputs with FLOPs reduced by $>$80\% compared to baseline models at 8,192 frames while maintaining comparable inference speed.

\input{tables/ablation_study}

\subsection{Discussions}

In this section, we conduct a comprehensive investigation to validate our design choices. We first evaluate the effectiveness of \textsc{dks} and \textsc{dfm}, and then analyze the sensitivity of model performance to the number of keyframes. Appendix~\ref{appendix:hyperparameters} further discusses other configuration settings.

\textbf{Ablation Study.}
We evaluate the effectiveness of \textsc{dks} and \textsc{dfm} through comparative experiments. For frame selection, we compare \textsc{dks} with two baselines: (1) query-guided sampling, which selects 32 frames with the highest query relevance scores (Eq.~\ref{dks_relevance}) without considering inter-frame redundancy, and (2) uniform sampling, which extracts 32 frames at fixed intervals. For feature merging, we compare \textsc{dfm} against (1) Q-former~\cite{li2023blip}, which integrates 729 visual token sequences into 1 fixed-length embeddings through a learnable cross-attention module, and (2) mean-pooling, which averages token embeddings within each non-keyframe into one token. All models are trained on the same dataset as mentioned in Appendix~\ref{appendix:training}.


Results in Table~\ref{table:ablation_study} demonstrate the effectiveness of both components. For frame selection, \textsc{dks} shows particularly strong improvements on longer videos (MLVU and Video-MME), while uniform sampling becomes less effective with increased video length due to missing key information, and query-guided sampling underperforms due to locally concentrated frame selection. For feature merging, \textsc{dfm} consistently outperforms alternatives across all datasets by effectively preserving contextual information, whereas mean-pooling suffers from redundant information and Q-former is limited by fixed embedding constraints.

\begin{figure}[!t]
    \centering
    \includegraphics[width=0.35\textwidth]{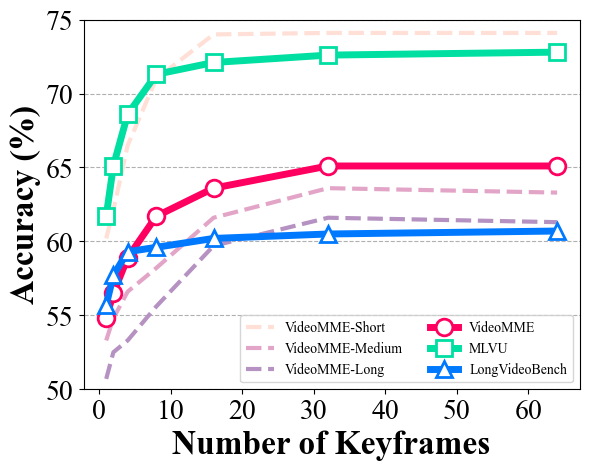}
    \caption{Analysis of model performance with varying numbers of keyframes across different video benchmarks. The test for Video-MME is conducted in non-subtitled setting.}
    \label{fig:keyframes}
\end{figure}

\textbf{Influence of Keyframes.} To determine the optimal trade-off between computational efficiency and model performance, we analyze \model's behavior with varying numbers of keyframes. Results in Figure~\ref{fig:keyframes} show consistent performance improvements as keyframe count increases, until reaching dataset-specific saturation points. Short-form videos (Video-MME-Short and LongVideoBench) plateau around 16 keyframes, while longer videos (Video-MME-Medium/Long and MLVU) require 32 keyframes for optimal performance. This aligns with our findings in \S~\ref{preliminary_fram} that informative frames can be sparsely sampled, though optimal sampling density varies with video length. 



\textbf{Effectiveness on other video tasks} Although \model's effectiveness has been verified on video question answering benchmarks, it is still unclear whether this method can truly preserve necessary information for more temporal-sensitive tasks like action recognition, video grounding or temporal reasoning. Therefore, we add some experiments.

For the video grounding task, we perform experiments on the widely used QVHighlights~\cite{lei2021detecting} benchmark. This dataset provides annotations for videos, marking the timestamps of query-relevant segments and the positions of keyframes. We evaluated \model~using the Hit@k metric, which measures the probability of the ground-truth keyframe appearing in the top-$k$ keyframes selected by our method. The results are as follows:

\input{tables/qvhighlights}

As shown in Table~\ref{table:qvhighlights}, \model~achieves 60\% accuracy for Hit@10 and 83\% for Hit@30, providing strong empirical support for its Dynamic Keyframe Selection (DKS) process. The current method effectively captures query-relevant video clips.

For the action recognition task, we incorporate two baselines: Adaframe~\cite{wu2019adaframe} and MGSampler~\cite{zhi2021mgsampler}, and evaluate them alongside \model~on three benchmarks: ActivityNet~\cite{caba2015activitynet}, Something-Something V2 (Sth-V2)~\cite{goyal2017something}, and UCF-101~\cite{soomro2012ucf101}. Since \model~is a generative model, we reformulate the original classification task as a multiple-choice problem during testing. The results are as follows:

\input{tables/action_recognition}

Notably, although \model~is not trained on these datasets, it demonstrates robust performance. This may be attributed to the generalizable knowledge acquired by the large-scale pretrained model, enabling effective zero-shot transfer to action recognition tasks.

%% file: tables/main_results.tex
\begin{table*}[!ht]
\small
\centering
\caption{Accuracy~(\%) on four video understanding benchmarks. \textbf{Size} indicates the number of parameters. \textbf{Frames} denotes either the fixed number of frames sampled from each video (e.g., 32 means sampling 32 frames regardless of video length) or the frame sampling rate (e.g., 2 FPS means sampling 2 frames per second). The best and second best results among open-source models of similar size (7$\sim$9B) are in \textbf{bold} and \underline{underlined}, respectively. For Video-MME, we report the performance on the whole benchmark~(``Overall'') and the long-form video subset (``Long'', videos $>$ 39 minutes). All baseline results are collected from their original papers. ``-'' indicates results not found.}
\label{table:main_result}
\begin{adjustbox}{max width=\textwidth}
\begin{tabular}{lrrccccc}
\toprule
\multicolumn{1}{c}{} & \multicolumn{1}{c}{} &  & \textbf{LVBench} & \multicolumn{1}{l}{\textbf{EgoSchema}} & \multicolumn{1}{l}{\textbf{LongVideoBench}} & \multicolumn{2}{c}{\textbf{Video-MME (wo / w sub)}} \\ 
\cmidrule(l){7-8} 
\multicolumn{1}{l}{\multirow{-2}{*}{\textbf{Models}}} & \multicolumn{1}{c}{\multirow{-2}{*}{\textbf{Size}}} & \multirow{-2}{*}{\textbf{Frames}} & \textbf{Overall} & \textbf{Validation} & \textbf{Validation} & \textbf{Overall} & \textbf{Long} \\
\midrule
\rowcolor[HTML]{EFEFEF} 
\multicolumn{1}{l}{\textbf{Average Duration}}
 & \multicolumn{1}{c}{\cellcolor[HTML]{EFEFEF}} &   & 4101s & 180s & 473s & 1,010s & \multicolumn{1}{c}{2,386s} \\ 
 \midrule
 \midrule
\rowcolor[HTML]{DDBBF2}
\multicolumn{8}{c}{\textit{\textbf{Proprietary Vision-Language Models}}} \\ \midrule
GPT4-V~\cite{openai2023gpt4}  & \multicolumn{2}{c}{\textit{Undisclosed}} & - & - & \textit{59.1} & \textit{59.9 / 63.3} & \textit{53.5 / 56.9} \\
GPT4-o~\cite{openaigpt4o} &\multicolumn{2}{c}{\textit{Undisclosed}} & \textit{30.8} & - & \textit{66.7} & \textit{71.9 / 77.2} & \textit{65.3 / 72.1} \\
Gemini-1.5-Pro~\cite{reid2024gemini} &\multicolumn{2}{c}{\textit{Undisclosed}} & \textit{33.1} & - & \textit{64.0} & \textit{75.0 / 81.3} & \textit{67.4 / 77.4} \\ 
\midrule
\midrule
\rowcolor[HTML]{DDBBF2}
\multicolumn{8}{c}{\textit{\textbf{Open-Source Multi-Image Vision-Language Models}}} \\
\midrule
\textcolor{gray}{LLaVA-OneVision~\cite{li2024llava}} & \textcolor{gray}{72B} & \textcolor{gray}{32} & \textcolor{gray}{-} & \textcolor{gray}{62.0} & \textcolor{gray}{61.3} & \textcolor{gray}{66.3 / 69.6} & \textcolor{gray}{60.0 / 62.4} \\
\textcolor{gray}{InternVL2~\cite{chen2024internvl}} & \textcolor{gray}{76B} & \textcolor{gray}{16} & \textcolor{gray}{-} & \textcolor{gray}{-} & \textcolor{gray}{61.0} & \textcolor{gray}{61.2 / 67.8} & \textcolor{gray}{-} \\
\midrule
LLaVA-OneVision~\cite{li2024llavaonevisioneasyvisualtask} & 7B & 32 & - & 60.1 & 56.5 & 58.2 / \ \ \ - \ \ \  & - \\
Oryx-1.5~\cite{liu2024oryx} & 7B & 128 & - & - & 56.3 & 58.8 / 64.2 & -  \\
InternVL2~\cite{chen2024internvl} & 8B & 16 & - & - & 54.6 & 56.3 / 59.3 & -  \\
InternVL2.5~\cite{chen2024expanding} & 8B & 64 & - & - & \underline{60.0} & \underline{64.2} / 66.9 & -  \\
Chat-Univi~\cite{jin2024chat} & 7B & 64 & - & - & - & 40.6 / 45.9 & 35.8 / 41.8  \\
MiniCPM-v2.6~\cite{yao2024minicpm} & 8B & 64 & - & - & 54.9 & 60.9 / 63.7 & 51.8 / 56.3 \\
mPLUG-Owl3~\cite{ye2024mplug} & 7B & 16 & - & - & 52.1 & 59.3 / \ \ \ - \ \ \  & 50.1 / \ \ \ - \ \ \  \\
Qwen2-VL~\cite{yang2024qwen2} & 7B & 2FPS & - & - & 55.6 & 63.3 / 69.0 & -  \\
NVILA~\cite{liu2024nvilaefficientfrontiervisual} & 7B & 256 & - & - & - & \underline{64.2} / \underline{70.0} & \underline{54.8} / \underline{63.3} \\ 
\midrule
\midrule
\rowcolor[HTML]{DDBBF2}
\multicolumn{8}{c}{\textit{\textbf{Open-Source Video-Language Models}}} \\
\midrule
\textcolor{gray}{VideoLLaMA2~\cite{cheng2024videollama}} & \textcolor{gray}{72B} & \textcolor{gray}{16} & \textcolor{gray}{-} & \textcolor{gray}{63.9} & \textcolor{gray}{-} & \textcolor{gray}{62.4 / 64.7} & \textcolor{gray}{57.6 / 59.0} \\
\midrule
LLaVA-Video~\cite{zhang2024video} & 7B & 1FPS & - & 65.6 & 58.2 & 63.3 / 69.7 & -  \\
LLaMA-VID~\cite{li2024llamavid} & 7B & 1FPS & \underline{23.9} & 38.5 & - & 25.9 / \ \ \ - \ \ \  & -  \\
Video-XL~\cite{shu2024video} & 7B & 2,048 & - & - & 49.5 & 55.5 / 61.0 & 49.2 / \ \ \ - \ \ \  \\
Video-LLaVA~\cite{lin2023video} & 7B & 8 & - & 38.4 & 37.6 & 39.9 / 41.6 & 36.2 / 38.1 \\
VideoLLaMA2~\cite{damonlpsg2023videollama} & 7B & 16 & - & 51.7 & - & 47.9 / 50.3 & - \\
VideoChat2~\cite{2023videochat} & 7B & 16 & - & 54.4 & 36.0 & 54.6 / \ \ \ - \ \ \  & 39.2 / \ \ \ - \ \ \  \\
Video-CCAM~\cite{fei2024videoccamenhancingvideolanguageunderstanding} & 9B & 96 & - & - & - & 53.2 / 57.4 & 46.7 / 49.9 \\
PLLaVA~\cite{xu2024pllava} & 7B & 16 & - & - & 40.2 & - & -  \\
Movie-Chat~\cite{song2024moviechat} & 7B & 2,048 & 22.5 & - & - & - & -  \\
Kangaroo~\cite{liu2024kangaroopowerfulvideolanguagemodel} & 8B & 64 & - & 62.7 & 54.8 & 56.0 / 57.6 & 46.7 / 59.3 \\
LongVU~\cite{shen2024longvu} & 7B & 1FPS & - & 67.6 & - & 60.6 / 59.5 & -  \\
LongVA~\cite{zhang2024longva} & 7B & 128 & - & - & - & 52.6 / 54.3 & 46.2 / 47.6 \\
LongVILA~\cite{chen2024longvilascalinglongcontextvisual} & 7B & 256 & - & \underline{67.7} & 57.1 & 60.1 / 65.1 & -  \\
\midrule
\rowcolor[HTML]{CFF2DF}
\multicolumn{1}{l}{\model (Ours)} & 7B & 1FPS & \textbf{46.3} & \textbf{74.2} & \textbf{60.2} & \textbf{67.7 / 71.6} & \multicolumn{1}{c}{\textbf{58.4 / 65.2}} \\
\bottomrule
\end{tabular}
\end{adjustbox}
\end{table*}

%% file: tables/ablation_study.tex
\begin{table}[!t]
\centering
\small
\caption{Ablation study results.}
\begin{adjustbox}{max width=\columnwidth}
\begin{tabular}{lccc}
\toprule
\multicolumn{1}{l}{\textbf{Model}} & \multicolumn{1}{c}{\textbf{LongVideoBnech}} & \multicolumn{1}{c}{\textbf{MLVU}} & \multicolumn{1}{c}{\textbf{Video-MME}} \\ \midrule
\multicolumn{1}{l}{\textbf{LLaVA-OneVision}} & 56.5 & 64.7 & 58.2 \\
\midrule
~~+\textbf{\textsc{dks} (Ours)} & 57.0 & 67.3 & 61.8 \\
~~+\textbf{Query-Guided Sampling} & 54.8 & 63.7 & 55.6 \\  
~~+\textbf{Uniformly Sampling} & 57.6 & 66.8 & 60.2 \\
\midrule
~~+\textbf{\textsc{dfm} (Ours)} & \textbf{60.5} & \textbf{72.6} & \textbf{65.1} \\ 
~~+\textbf{Q-former} & 53.9 & 62.6 & 59.2 \\
~~+\textbf{Mean-Pooling} & 56.3 & 67.3 & 62.1 \\
\bottomrule

\end{tabular}
\end{adjustbox}
\label{table:ablation_study}
\end{table}

%% file: tables/qvhighlights.tex
\begin{table}[htbp]
\centering
\caption{Evaluation results on QVHighlights.}
\label{table:qvhighlights}
\begin{tabular}{cccc}
\toprule
\textbf{Hit@1} & \textbf{Hit@5} & \textbf{Hit@10} & \textbf{Hit@30} \\ \midrule
26.1 & 47.3 & 60.0 & 83.2 \\ \bottomrule
\end{tabular}
\end{table}

%% file: tables/action_recognition.tex
\begin{table}[htbp]
\tiny
\centering
\caption{Evaluation on action recognition tasks. The results for Adaframe and MGSampler are sourced from their papers.}
\label{table:action_recognition}
\begin{tabular}{lccc}
\toprule
\textbf{Method} & \textbf{ActivityNet (mAP \%)} & \textbf{Sth-V2 (Acc. \%)} & \textbf{UCF-101 (Acc. \%)} \\ \midrule
Adaframe & 71.5 & - & - \\
MGSampler & - & 60.1 & 95.2 \\
\model (Ours) & \textbf{78.6} & \textbf{86.9} & \textbf{97.6} \\ \bottomrule
\end{tabular}
\end{table}

%% file: conclusion.tex
\section{Conclusion}
In this work, we address the critical challenge of scaling video-language models (VLMs) to process ultra-long videos efficiently. We present~\model, a video-language model guided by differential distillation. 
By combining Differential Keyframe Selection (\textsc{dfs}) and Differential Feature Merging (\textsc{dfm}), \model~demonstrate superior performance across four video benchmarks, particularly in long-form scenarios. 
On the challenging VideoNIAH benchmark, \model~processes videos up to 10K frames ($\sim$2.7 hours) on a single NVIDIA A100 GPU, outperforming state-of-the-art models by 12.82\% in accuracy while significantly reducing computational overhead. The hierarchical compression strategy enables stable performance across varying video lengths, with minimal memory growth and latency.
Our work establishes a scalable paradigm for long-form video understanding, bridging the gap between computational constraints and real-world demands.

%% file: appendix.tex
\section{Experiment Details}
\subsection{Benchmarks}
\label{appendix:benchmarks}
\paragraph{LVBench~\cite{wang2024lvbench}} consists of publicly sourced long videos with an average duration of approximately 4,101 seconds, significantly longer than existing datasets. The videos are categorized into six major categories and 21 subcategories, covering a wide range of topics such as sports, live streams, TV shows, documentaries, and animations. The dataset is annotated through a combination of manual effort and model assistance. 
\paragraph{EgoSchema~\cite{mangalam2023egoschema}} is a diagnostic benchmark aimed at assessing the long-form video-language understanding capabilities of modern multimodal systems. Derived from the Ego4D dataset, EgoSchema consists of over 5000 human-curated multiple-choice question-answer pairs, covering more than 250 hours of high-quality egocentric video data (180 seconds in average). The dataset is designed to require models to understand and reason over extended temporal contexts, with each question based on a three-minute video clip.
\paragraph{LongVideoBench~\cite{wu2024longvideobench}} comprises 3,763 videos with varying lengths, ranging from 8 seconds to 1 hour, and 6,678 human-annotated multiple-choice questions. The videos cover a wide range of themes, including movies, news, life and knowledge. The benchmark introduces a novel task called "referring reasoning" to address the single-frame bias in existing video understanding metrics. This task requires models to process more frames to improve performance.
\paragraph{VideoMME~\cite{fu2024video}} consists of 900 videos spanning 6 primary visual domains (Knowledge, Film \& Television, Sports Competition, Artistic Performance, Life Record, and Multilingual) with 30 subfields. The videos vary in length from 11 seconds to 1 hour, covering short, medium, and long durations. Each video is annotated with 3 multiple-choice questions, resulting in a total of 2,700 question-answer pairs. It also includes subtitles and audio tracks for 744 and 900 videos. 
\subsection{Baselines}
\label{appendix:baselines}
\subsubsection{Proprietary Vision-Language Models}
\paragraph{GPT4-V~\cite{openai2023gpt4}} is an advanced multimodal AI model developed by OpenAI, combining the power of natural language processing with visual understanding. Building on the capabilities of GPT-4, GPT-4-V integrates vision-based functionalities, enabling it to generate responses based on both text and visual inputs.
\paragraph{GPT4-o~\cite{openaigpt4o}} is the latest iteration in the Generative Pre-trained Transformer series. Building on the strengths of its predecessors, GPT-4-o offers enhanced capabilities in understanding multimodal inputs.
\paragraph{Gemini-1.5-Pro~\cite{reid2024gemini}} is Google's latest breakthrough in artificial intelligence, representing a significant leap in multi-modal and long-context understanding.

\subsubsection{Open-Source Multi-Image Vision-Language Models}
\paragraph{LLaVA-OneVision~\cite{li2024llava}} leverages strong transfer learning across different modalities, pushing performance boundaries in tasks such as video understanding and complex reasoning through a unified model architecture.
\paragraph{Oryx-1.5~\cite{liu2024oryx}} is a novel multi-modal large language model (MLLM) designed for on-demand spatial-temporal understanding of diverse visual inputs, including images, videos, and 3D scenes. It processes visual inputs at arbitrary resolutions and temporal lengths through a pre-trained OryxViT model and a dynamic compressor module.
\paragraph{InternVL Series~\cite{chen2024internvl,chen2024expanding}} is characterized by its large-scale visual encoder, InternViT-6B, and language middleware, QLLaMA, which enables it to handle a wide range of vision-language tasks. InternVL 2.5 is the latest version in the InternVL series, and it significantly enhances its performance through improvements in training and testing strategies, as well as data quality. The model demonstrates outstanding performance across multiple benchmarks.
\paragraph{Chat-Univi~\cite{jin2024chat}} uses of dynamic visual tokens to represent images and videos in a unified framework, allowing it to capture spatial details of images and temporal relationships of videos efficiently. Additionally, the model employs a multi-scale representation that enables it to perceive both high-level semantic concepts and low-level visual details.
\paragraph{MiniCPM-v2.6~\cite{yao2024minicpm}} achieves strong performance comparable to GPT-4V while being optimized for efficiency. It has high-resolution image perception, strong OCR capabilities, multilingual support for over 30 languages, and low hallucination rates, making it suitable for real-world applications with limited computational resources.
\paragraph{mPLUG-Owl3~\cite{ye2024mplug}} is a versatile multi-modal large language model designed to enhance long image-sequence understanding. It introduces Hyper Attention blocks, which integrate vision and language efficiently by allowing parallel cross-attention and self-attention within the transformer architecture.
\paragraph{Qwen2-VL~\cite{Qwen2VL}} is an advanced vision-language model that enhances the perception of visual information at any resolution. Qwen2-VL introduces Naive Dynamic Resolution mechanism, which allows the model to process images of varying resolutions into different numbers of visual tokens, and the Multimodal Rotary Position Embedding (M-RoPE), which effectively fuses positional information across text, images, and videos.
\paragraph{NVILA~\cite{liu2024nvilaefficientfrontiervisual}} builds on VILA and enhances its architecture by scaling up spatial and temporal resolutions before compressing visual tokens, enabling efficient processing of high-resolution images and long videos. Additionally, NVILA incorporates system-wide optimizations, including dataset pruning, FP8 training, and specialized inference engines, to achieve significant improvements in training speed, memory usage, and inference efficiency, while maintaining competitive performance across various benchmarks.

\subsubsection{Open-Source Video Vision-Language Models}
\paragraph{LLaVA-Video~\cite{zhang2024video}} is trained on a high-quality synthetic dataset LLaVA-Video-178K. It processes frames using an optimized video representation technique called LLaVA-Video SlowFast, which allows for more detailed temporal understanding compared to previous models.
\paragraph{LLaMA-VID~\cite{li2024llamavid}} represents each frame with two distinct tokens—a context token that captures the overall image context based on user input, and a content token that encapsulates visual details. This dual-token strategy significantly reduces the computational burden of processing long videos while preserving critical information, enabling VLMs to handle hour-long videos.
\paragraph{Video-XL~\cite{shu2024video}} leverages the inherent key-value sparsification capability of large language models to condense visual inputs through a new special token called the Visual Summarization Token (VST). This approach enables the model to handle long videos efficiently by summarizing visual information into compact representations, while dynamic compression and curriculum learning strategies further enhance its performance and training effectiveness.
\paragraph{Video-LLaVA~\cite{lin2023video}} uses a LanguageBind encoder to align visual signals from images and videos into a unified visual feature space, followed by joint training of images and videos to enhance multi-modal interactions.
\paragraph{VideoLLaMA2~\cite{cheng2024videollama}} is designed to enhance spatial-temporal modeling and audio understanding in video and audio-oriented tasks. It introduces a Spatial-Temporal Convolution (STC) connector that effectively captures the intricate spatial and temporal dynamics of video data, and an integrated Audio Branch that enriches the model’s multimodal understanding capabilities by seamlessly incorporating audio cues.
\paragraph{VideoChat2~\cite{2023videochat}} utilizes the UMT-L visual encoder, which is specifically designed for spatiotemporal representation learning. It also adopt Q-former to compresses redundant visual tokens into fewer tokens and aligns them with text tokens. VideoChat2 excels in tasks that require understanding of temporal sequences, such as action prediction, action sequence, and moving direction.
\paragraph{Video-CCAM~\cite{fei2024videoccamenhancingvideolanguageunderstanding}} is designed to enhance video-language understanding for both short and long videos. It uses Causal Cross-Attention Masks (CCAMs) within the cross-attention layers, which allows the model to effectively process a large number of visual tokens while preserving temporal order in videos.
\paragraph{PLLaVA~\cite{xu2024pllava}} is a novel and efficient approach to adapt existing image-language pre-trained models for dense video understanding. PLLaVA employs a simple yet effective pooling strategy to smooth the feature distribution along the temporal dimension, reducing the impact of high-norm visual features that often lead to performance degradation. 
\paragraph{Movie-Chat~\cite{song2024moviechat}} introduces a innovative memory mechanism inspired by the Atkinson-Shiffrin memory model, which employs a combination of short-term and long-term memory to efficiently process and understand long videos.
\paragraph{Kangaroo~\cite{liu2024kangaroopowerfulvideolanguagemodel}} is designed for long-context video input. Kangaroo addresses the challenges of limited training data and excessive compression of visual features in long videos by developing a data curation system to build a large-scale high-quality dataset for vision-language pre-training and instruction tuning. It also employs a curriculum training pipeline that gradually increases the resolution and number of input frames to enhance the model's ability to process long videos.
\paragraph{LongVU~\cite{shen2024longvu}} introduces a spatiotemporal adaptive compression mechanism designed for long video-language understanding. LongVU leverages cross-modal queries and inter-frame dependencies to reduce the number of video tokens while preserving visual details, making it possible to process hour-long videos within the context length constraints of large language models (LLMs).
\paragraph{LongVA~\cite{zhang2024longva}} leverages a unique approach called "long context transfer," where it extends the context length of a language model by training it on longer text data and then aligns this extended model with visual inputs using short image data. This enables LongVA to process over 2,000 frames or more than 200K visual tokens without explicit long video training.
\paragraph{LongVILA~\cite{anonymous2024longvila}} adopts a five-stage training pipeline that incorporates multi-modal alignment, large-scale pre-training, supervised fine-tuning, context extension, and long video fine-tuning. Additionally, LongVILA utilizes a novel Multi-Modal Sequence Parallelism (MM-SP) system, which efficiently parallelizes training and inference for long videos.

\subsection{Implement Details}
\label{appendix:training}

\input{tables/training_datasets}

\paragraph{Training Datasets}

We employ a three-stage training strategy to fine-tune \model~and enhance its video comprehension capabilities. 
During the pretraining phase, the primary objective is to adapt \model~to the interleaved input structure combining keyframes and non-keyframes. Our training data comprises approximately 7.4M video-caption pairs collected from multiple open-source datasets including WebVid~\cite{Bain21}, InternVid~\cite{wang2023internvid}, ShareGPTVideo~\cite{zhang2024direct}, OpenVid~\cite{nan2024openvid}, and Vript~\cite{yang2024vript}, ensuring broad coverage of diverse visual scenarios.
This comprehensive collection trains the model to develop a foundational understanding of video inputs.

Additionally, we further expand the data scale and enhance the model's capabilities by incorporating diverse types of QA datasets, including VideoChat2-IT~\cite{2023videochat}, EgoTaskQA~\cite{jia2022egotaskqa}, CLEVRER~\cite{yi2019clevrer}, LLaVA-Video-178K~\cite{zhang2024video}, MovieChat~\cite{song2023moviechat}, PerceptionTest~\cite{patraucean2023perception}, HiREST~\cite{zala2023hierarchical}, STAR~\cite{wu2021benchmark}, and NExTQA~\cite{xiao2021next}, ultimately constructing a dataset of 1.3M samples. This dataset improves the model's instruction-following ability, enabling it to better adapt to various downstream tasks. Specifically, for LLaVA-Video-178K, we utilize the \textit{academic} and \textit{youtube} subsets.

Finally, to improve the model's performance on longer videos, we incorporate FineVideo~\cite{Farré2024FineVideo} and CinePile~\cite{rawal2024cinepile}, two datasets with longer average video durations (280s and 160s, respectively), and get approximately 500K training samples.

\paragraph{Training Parameters} 
As mentioned in \S~\ref{appendix:hyperparameters}, we configure $\alpha=10^{-2}$ and $\tau=0.85$ to achieve balanced effectiveness. 
During training, we set learning rate to $2\times10^{-6}$ for the visual encoder and $10^{-5}$ for the remaining components. We use adamW as the optimizer. The optimization follows a cosine learning rate scheduler with a warmup ratio of 0.03. We set the batch size to 1 and the gradient accumulation steps to 4. We train the model for 1 epoch, which costs approximately two weeks with 32 NVIDIA A100 GPUs.

\section{Discussions}

\subsection{Computational Efficiency Result}
\label{appendix:efficiency}
We compare against four representative video-language models: LongVA, LLaVA-Video, LLaMA-VID, and VideoChat-Flash. As shown in Table~\ref{table:efficiency},
\model~demonstrates a significant advantage in memory efficiency. Unlike typical models whose memory usage increases drastically with input length, \model~shows only a marginal increase in memory demand as the input length grows. This characteristic enables \model~to handle extremely long inputs under limited memory constraints, which can be attributed to the effective removal of redundant information through our differential distillation. 
Furthermore, \model~exhibits a slow growth rate of FLOPs with input expansion, demonstrating remarkable superiority in handling long sequences. When the input length reaches the thousand-frame level, \model~outperforms all baseline models, requiring merely 18.4\% of the FLOPs consumed by VideoChat-Flash when processing 8,192 frames.

Moreover, \model~also demonstrates a significant speed advantage, with TTFT results indicating that it exhibits a response latency that is merely 50\% or less of typical models. When the number of frames reaches 8,192, \model~exhibits marginally slower performance compared to VideoChat-Flash. This is attributed to the pruning operations implemented in the Large Language Model (LLM) component of VideoChat-Flash, which specifically optimizes inference speed. Nevertheless, \model~achieves comparable performance levels without employing such optimizations.

\input{tables/efficiency}

\subsection{Influence of Hyperparameter Configuration}
\label{appendix:hyperparameters}

We conduct a series of experiments under various settings of $\alpha$ and $\tau$. As reported in Table~\ref{table:hyperparameters}, the overall performance of the model first improves and then decreases as $\tau$ increases. This trend can be attributed to the fact that a too small $\tau$ leads to insufficient sampling, while an excessively large $\tau$ results in overly concentrated sampling, both of which negatively impact the model's effectiveness. We find that setting $\tau=0.85$ provides a robust balance between semantic preservation and temporal diversity. Further analysis of $\alpha$ reveals a similar pattern. To some extent, reducing $\alpha$ allows the weight distribution to focus more on the target tokens, which helps in eliminating redundant information. However, when $\alpha$ is too small, the weights concentrate on only a few tokens, leading to information loss that adversely affects the model's performance. Through extensive experimentation, we find that setting $\alpha=10^{-2}$ ensures robust performance across diverse video content.

\input{tables/hyperparameters}

%% file: tables/training_datasets.tex
\begin{table}[htbp]
\begin{tabular}{lp{11cm}l}
\toprule
\multicolumn{1}{c}{\textbf{Stage}} & \multicolumn{1}{c}{\textbf{Datasets}} & \multicolumn{1}{c}{\textbf{Size}} \\ \midrule
\rowcolor[HTML]{EFEFEF} 
Pretrain & WebVid~\cite{Bain21}, InternVid~\cite{wang2023internvid}, ShareGPTVideo~\cite{zhang2024direct}, OpenVid~\cite{nan2024openvid}, Vript~\cite{yang2024vript} & 7.4M \\
Short Video Understanding & VideoChat2-IT~\cite{2023videochat}, EgoTaskQA~\cite{jia2022egotaskqa}, CLEVRER~\cite{yi2019clevrer}, LLaVA-Video-178K~\cite{zhang2024video}, MovieChat~\cite{song2023moviechat}, PerceptionTest~\cite{patraucean2023perception},  STAR~\cite{wu2021benchmark}, NExTQA~\cite{xiao2021next} & 1.3M \\
\rowcolor[HTML]{EFEFEF} 
Longer Video Understanding & FineVideo~\cite{Farré2024FineVideo}, CinePile~\cite{rawal2024cinepile} & 500K \\ \bottomrule
\end{tabular}
\end{table}

%% file: tables/efficiency.tex
\begin{table}[h]
\centering
\caption{Computational efficiency comparison across different input lengths.}
\begin{adjustbox}{max width=\columnwidth}
\begin{tabular}{lrcc}
\toprule
\textbf{Model} & \textbf{FLOPs~(T)} & \textbf{Memory~(MB)} & \textbf{TTFT~(ms)} \\
\hline
\hline
\multicolumn{4}{c}{\cellcolor[HTML]{EFEFEF}\textit{Input 64 Frames}} \\ \hline
\textbf{LongVA} & 316.0 & 62,262 & 798 \\
\textbf{LLaVA-Video} & 101.8 & 32,362 & 203 \\
\textbf{LLaMA-VID} & 112.3 & 46,402 & 373 \\
\textbf{VideoChat-Flash} & \textbf{29.6} & \textbf{18,688} & 521 \\
\textbf{\model} & 61.0 & 23,201 & \textbf{80} \\ \hline
\hline
\multicolumn{4}{c}{\cellcolor[HTML]{EFEFEF}\textit{Input 256 Frames}} \\ \hline
\textbf{LongVA} & 2,946.3 & \textit{OOM} & \textit{OOM} \\
\textbf{LLaVA-Video} & 1,221.6 & 71,495 & 535 \\
\textbf{LLaMA-VID} & 245.9 & 47,294 & 492 \\
\textbf{VideoChat-Flash} & \textbf{113.3} & \textbf{22,665} & 586 \\
\textbf{\model} & 120.3 & 23,647 & \textbf{157} \\ 
\hline
\hline
\multicolumn{4}{c}{\cellcolor[HTML]{EFEFEF}\textit{Input 1,024 Frames}} \\ \hline
\textbf{LongVA} & 27,605.6 & \textit{OOM} & \textit{OOM} \\
\textbf{LLaVA-Video} & 18,324.0 & \textit{OOM} & \textit{OOM} \\
\textbf{LLaMA-VID} & 1,045.3 & 48,678 & 785 \\
\textbf{VideoChat-Flash} & 489.6 & 38,427 & 694 \\
\textbf{\model} & \textbf{356.8} & \textbf{24,339} & \textbf{462} \\ \hline
\hline
\multicolumn{4}{c}{\cellcolor[HTML]{EFEFEF}\textit{Input 8,192 Frames}} \\ \hline
\textbf{LongVA} & 2,153,901.7 & \textit{OOM} & \textit{OOM} \\
\textbf{LLaVA-Video} & 1,643,892.3 & \textit{OOM} & \textit{OOM} \\
\textbf{LLaMA-VID} & 20,906.0 & 91,438 & 5,080 \\
\textbf{VideoChat-Flash} & 13,896.6 & 92,034 & \textbf{2,730} \\
\textbf{\model} & \textbf{2,565.2} & \textbf{45,819} & 3,374 \\ \bottomrule
\end{tabular}
\end{adjustbox}
\label{table:efficiency}
\end{table}

%% file: tables/hyperparameters.tex
\begin{table}[htbp]
\centering
\caption{Performance varying with hyperparameters $\alpha$ and $\tau$.}
\label{table:hyperparameters}
\begin{tabular}{cr|ccc}
\toprule
\textbf{$\alpha$} & \multicolumn{1}{c}{\textbf{$\tau$}} & \textbf{LongVideoBench} & \textbf{MLVU} & \textbf{Video-MME} \\ \midrule
\multirow{4}{*}{\textbf{$1$}} & 0.35 & 55.6 & 65.9 & 59.4 \\
 & 0.50 & 58.3 & 69.5 & 61.2 \\
 & 0.85 & \textbf{58.9} & \textbf{70.5} & \textbf{62.7} \\
 & 1.00 & 51.3 & 64.5 & 58.0 \\ \midrule
\multirow{4}{*}{\textbf{$10^{-1}$}} & 0.35 & 57.2 & 63.3 & 60.5 \\
 & 0.50 & 58.0 & 68.9 & 62.6 \\
 & 0.85 & \textbf{59.1} & \textbf{71.5} & \textbf{64.5} \\
 & 1.00 & 51.1 & 63.6 & 57.3 \\ \midrule
\multirow{4}{*}{\textbf{$10^{-2}$}} & 0.35 & 56.8 & 65.3 & 60.2 \\
 & 0.50 & 59.6 & 69.7 & 63.2 \\
 & 0.85 & \textbf{60.5} & \textbf{72.6} & \textbf{65.1} \\
 & 1.00 & 50.9 & 62.3 & 58.8 \\ \midrule
\multirow{4}{*}{\textbf{$10^{-3}$}} & 0.35 & 49.3 & 61.6 & 45.6 \\
 & 0.50 & 52.2 & 63.4 & 47.0 \\
 & 0.85 & \textbf{54.4} & \textbf{65.6} & \textbf{48.2} \\
 & 1.00 & 46.5 & 60.9 & 43.6 \\ \bottomrule
\end{tabular}
\end{table}